\documentclass[12pt]{article}

\usepackage{setspace}
\usepackage{natbib}
\usepackage{graphicx}
\usepackage{caption}
\usepackage{subcaption}
\usepackage{graphics}
\usepackage{hyperref}
\usepackage{url}
\usepackage{etoolbox}
\usepackage{bm}
\usepackage{amsmath,amssymb,amsthm}
\usepackage[linesnumbered,ruled,vlined]{algorithm2e}
\usepackage{float}
\usepackage{multirow}
\usepackage{colortbl}
\usepackage{enumerate}
\usepackage{adjustbox}

\newcommand{\blind}{0}

\newcommand{\xred}{\boldsymbol{X}_R}
\newcommand{\xadd}{\boldsymbol{X}_A}

\newcommand{\predpoints}{\boldsymbol{x}_{\text{\tiny TEST}}}

\definecolor{darkgreen}{rgb}{0.0, 0.5, 0.0}

\begin{document}

\def\spacingset#1{\renewcommand{\baselinestretch}%
{#1}\small\normalsize} \spacingset{1}

\if0\blind
{
  \title{\bf Formal Hypothesis Tests for Additive Structure in Random Forests}
  \author{Lucas Mentch\thanks{
     We would like to thank Cornell University's Lab of Ornithology for providing interesting data.  Giles Hooker was partially supported by NSF grants DMS-1053252 and DEB-1353039 and NIH grant R03DA036683.}\hspace{.2cm}\\
    and \\
    Giles Hooker \\
    Department of Statistical Science \\ Cornell University}
  \maketitle
} \fi

\if1\blind
{
  \bigskip
  \bigskip
  \bigskip
  \begin{center}
    {\LARGE\bf Formal Hypothesis Tests for Additive Structure in Random Forests}
\end{center}
  \medskip
} \fi

\bigskip
\begin{abstract}
\noindent While statistical learning methods have proved powerful tools for predictive modeling, the black-box nature of the models they produce can severely limit their interpretability and the ability to conduct formal inference.  However, the natural structure of ensemble learners like bagged trees and random forests has been shown to admit desirable asymptotic properties when base learners are built with proper subsamples.  In this work, we demonstrate that by defining an appropriate grid structure on the covariate space, we may carry out formal hypothesis tests for both variable importance and underlying additive model structure.  To our knowledge, these tests represent the first statistical tools for investigating the underlying regression structure in a context such as random forests.  We develop notions of total and partial additivity and further demonstrate that testing can be carried out at no additional computational cost by estimating the variance within the process of constructing the ensemble.  Furthermore, we propose a novel extension of these testing procedures utilizing random projections in order to allow for computationally efficient testing procedures that retain high power even when the grid size is much larger than that of the training set.
\end{abstract}

\vfill

\newpage
\spacingset{1.45} 


\section{Introduction}
As scientific data grows larger and becomes easier to collect, traditional statistical models often prove insufficient for fully capturing the underlying process.  Learning algorithms, on the other hand, adapt well to a variety of data types and produce accurate predictions, but their inherent complexity and black-box nature makes addressing even the simplest scientific questions significantly more difficult.  This work provides a formal statistical test for determining variable interactions whenever ensemble learning methods like random forests are used as the primary modeling tool.

Additive models were suggested by \cite{Friedman1981} and further developed and made popular by \cite{Stone1985} and \cite{HastieAdditiveModels}.  An underlying regression function $F\colon \mathcal{X} \mapsto \mathbb{R}$ is said to be additive if
\[
F(x_1, ..., x_d) = \sum_{i=1}^{d} F_{i}(x_i)
\]
\noindent for some functions $F_1, ..., F_d$.  If the regression function cannot be written as, or at least well-approximated by, a sum of univariate functions, then an interaction exists between some subset of the covariates.  Many methods have been developed to estimate the additive functions $F_1, ..., F_d$ including a method based on marginal integration by \cite{Linton1995}, a wavelet method suggested by \cite{Amato2001}, a tree-based method by \cite{Lou2013}, and the most popular class based on backfitting algorithms as found in \cite{Buja1989}, \cite{Opsomer1998,Opsomer1999}, and \cite{Mammen1999}.

The popularity of additive models and their ease of interpretation has inspired hypothesis tests to assess whether observed data should be modeled in an additive fashion.  Versions of these lack-of-fit tests have been proposed by \cite{Barry1993}, \cite{Eubank1995}, \cite{Dette2001ANOVA}, \cite{Derbort2002}, and \cite{De2004}.  \cite{FanAdditivity} further extend these procedures to also evaluate whether the additive components belong to a particular parametric class.  Even when additive models are not used as the primary analytical tool, scientists often utilize these and related interaction detection methods to determine which variables contribute additively to the response; when no interactions are detected, the levels of one feature may be changed without affecting the contribution to the response made by the others.


Their utility notwithstanding, additive models can often fail to fully capture the signal hidden within modern complex data, even when relatively little signal results from variable interactions.  On the other hand, learning algorithms like bagged trees and random forests introduced by \cite{bagging, randomforests}, are robust to a variety of regression functions and are considered something of a gold standard in terms of predictive accuracy.  Though this accuracy continues to drive their popularity, little is understood about the underlying mathematical and statistical properties of these ensemble methods.  Thus, while practitioners routinely rely on such methods to make predictions, when standard results such as confidence intervals or p-values from hypothesis tests for variable importance or interactions need reported, those practitioners are forced to move to an entirely different modeling technique and rely on more well-established procedures.  At best, the ensembles might be used to better inform which hypotheses to test and/or which variables should be included in a simpler model.

Recently however, important progress has been made in understanding the asymptotic properties of these ensemble methods by considering a subsampling approach in lieu of the traditional bootstrapping procedure.  \cite{MHJMLR} show that when proper subsamples are used to construct individual trees, the ensemble predictions can be seen as extensions of classical U-statistics and as such, are asymptotically normal.  \cite{WagerIJ} apply recent results on the infinitesimal jackknife \citep{Efron2014} to produce estimates of standard errors for subsampled random forest predictions and \cite{Wager2015} later demonstrate the consistency of such an approach.  Most recently, \cite{Scornet2015} provided the first consistency results for Breiman's original random forest procedure when subsampling is employed and the underlying regression function has an additive form.

This paper continues in this recent trend by developing formal hypothesis tests for additivity in ensemble learners like bagged trees and random forests.  These tests allow practitioners to formally investigate the manner in which features contribute to the response when simpler, more direct tools are insufficient and to our knowledge, represent the first formal procedures for investigating the structure of the underlying regression function within the context of ensemble learning.  That is, statistically valid results such as p-values may be gathered directly from the ensemble instead of relying on \emph{ad hoc} measures or appealing to a simplified model.  In Section \ref{sec:2} we propose a formal test for feature significance by imposing a grid structure on the covariate space and in Section \ref{sec:3} we demonstrate that this additional structure further allows for tests of additivity.  In Section \ref{sec:4} we incorporate random projections to extend our procedure to the situation where a large test grid is needed, so as to accommodate potential high dimensional settings.  Finally, in Sections \ref{sec:5} and \ref{sec:6}, we provide simulations to investigate the power of our hypothesis tests and apply our testing procedures to an ecological dataset.

\section{Hypothesis tests for feature significance}
\label{sec:2}

Recent theory has demonstrated that a subsampling approach to constructing supervised ensembles like random forests may allow these learners to be reigned in within the realm of traditional statistical inference.  Specifically, \cite{MHJMLR} show that by controlling the subsample growth rate, individual predictions are asymptotically normal thereby paving the way for a formal method of evaluating variable (feature) significance.  As a simple example, consider a setting with just two features $X_1$ and $X_2$ where the response observed according to $Y = F(X_1, X_2) + \epsilon$.  To test the significance of $X_2$, we can generate a test set $\predpoints$ consisting of $N$ points and build two subsampled ensembles $\hat{F}$ and $\hat{F}_1$.  Both ensembles employ the same subsamples, but $\hat{F}$ is constructed using both $X_1$ and $X_2$ whereas $\hat{F}_1$ is built using only $X_1$.  Predictions at each point in $\predpoints$ are then made with each ensemble and \cite{MHJMLR} show that the vector of differences in predictions has a multivariate normal limiting distribution with mean $\mu$ and variance $\Sigma$.  Given consistent estimators of these parameters, $\hat{\mu}^{T} \hat{\Sigma}^{-1} \hat{\mu} \sim \chi_{N}^2$ can be used as a test statistic to formally evaluate the hypotheses
\begin{align}
\label{hyp:Sig2feat} &H_0:  F(x_1, x_2) = F_1(x_1) \; \; \forall (x_1, x_2) \in \predpoints \\
&H_1:  F(x_1, x_2) \neq F_1(x_1) \; \; \mbox{for some } (x_1, x_2) \in \predpoints \; \; \mbox{for any } F_1. \notag
\end{align}

\noindent Though asymptotically valid, this procedure requires building separate ensembles for each feature of interest.  We demonstrate here that imposing additional structure on the test set allows us to both avoid training an additional set of trees and also perform tests for additivity.

\begin{figure}
  \centering
  \includegraphics[scale=0.4]{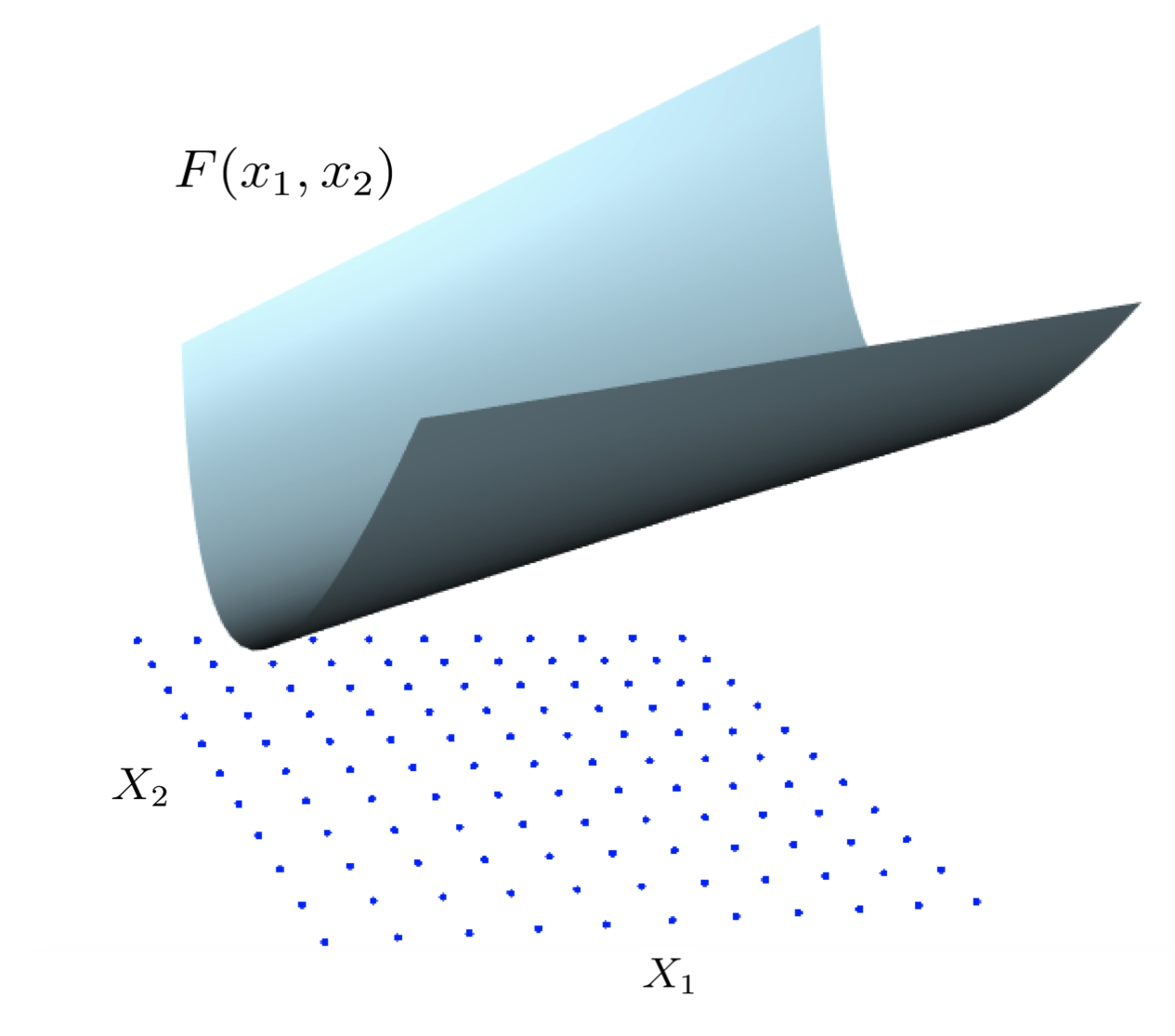}
  \caption{\label{fig:grid} A grid of test points shown in the $X_1$$X_2$ plane below the response surface.}
\end{figure}

Define a grid consisting of $N$ total test points as in Figure \ref{fig:grid} with $N_1$ levels $x_{1_i}$ and $N_2$ levels $x_{2_j}$ so that the $(i,j)^{th}$ point in the grid has true value $F_{ij} = F(x_{1_i},x_{2_j})$ and predicted value $\hat{F}_{ij}$.  In the case of categorical covariates, these grid levels are naturally occurring while in the case of continuous covariates, these levels can be specified as appropriate (e.g.\ based on quantiles of the observed data).  Let $V_F$ and $V_{\hat{F}}$ represent the vectorized versions of these true and predicted values so that $V_{F} = (F_{1,1}, ..., F_{1,N_2}, ..., F_{N_1,1}, ..., F_{N_1, N_2})^T$ and define
\[
\hat{f}_{i \cdot } = \frac{1}{N_2} \sum_{j=1}^{N_2} \hat{F}_{ij}
\]
\noindent as the average response at the $i^{th}$ level $x_{1_i}$ across all grid levels $x_{2_j}$.  For each point in the grid, the difference in predictions $\hat{F}_{ij} - \hat{f}_{i \cdot}$ can be written in vectorized form as $DV_{\hat{F}}$ for an $N \times N$ difference matrix $D$ of rank $N-N_1$.  In this case, $D = I_{N} - \left( I_{N_1} \otimes \frac{1}{N_2} \bm{1}_{N_2 \times N_2} \right)$ where $I_C$ is the $C \times C$ identity matrix, $\bm{1}_{C \times C}$ is the $C \times C$ matrix of 1's, and $\otimes$ denotes the standard tensor product.  Let $\Sigma$ denote the covariance of $V_F$ and $\hat{\Sigma}$ a consistent covariance estimate of the predictions.  Then we can define $\Sigma_D = cov(DV_F) = D \Sigma D^T$ so that $\hat{\Sigma}_{D} = D \hat{\Sigma} D^{T}$ forms a consistent estimate of the covariance of the projected predictions $\Sigma_{D}$.  Then $(DV_{\hat{F}})^{T} \hat{\Sigma}^{-1}_{D} DV_{\hat{F}} \sim \chi_{N-N_1}^{2}$ and since we can equivalently write the hypotheses in (\ref{hyp:Sig2feat}) as
\begin{align}
&H_0:  F_{ij} - f_{i \cdot} = 0 \; \; \forall (x_1,x_2) \in \predpoints \notag \\
&H_1:  F_{ij} - f_{i \cdot} \neq 0 \; \; \mbox{for some } (x_1, x_2) \in \predpoints \notag
\end{align}

\noindent $(DV_{\hat{F}})^{T} \hat{\Sigma}^{-1}_{D} DV_{\hat{F}}$ can be used as a test statistic.

%
%

Asymptotically, this test statistic has a $\chi_{N-N_1}^{2}$ distribution and thus can be compared to the $1-\alpha$ quantile to achieve a test with type 1 error rate $\alpha$; if the test statistic is larger than this critical value, we reject the null hypothesis and conclude that $X_2$ is significant.

This testing procedure readily extends to the more general case of $d$ features $X_1, ..., X_d$.  Let $\xred$ and $\xadd$ form a partion of $\{ X_1, ..., X_d \}$ so that $\xred$ and $\xadd$ are disjoint and $\xred \cup \xadd = \{ X_1, ..., X_d \}$; the set $\xred$ denotes the \emph{reduced} set of features and $\xadd$ represents the \emph{additional} features that we want to test for significance.  To test the hypotheses
\begin{align*}
&H_0:  F(\boldsymbol{x}_{R_i}, \boldsymbol{x}_{A_i}) = F_R(\boldsymbol{x}_{R_i}) \; \; \forall (\boldsymbol{x}_{R_i}, \boldsymbol{x}_{A_i}) \in \predpoints \\
&H_1:  F(\boldsymbol{x}_{R_i}, \boldsymbol{x}_{A_i}) \neq F_R(\boldsymbol{x}_{R_i}) \mbox{ for some } (\boldsymbol{x}_{R_i}, \boldsymbol{x}_{A_i}) \in \predpoints \mbox{ for any } F_R
\end{align*}

\noindent we simply repeat the testing procedure in the above example, replacing the levels $x_{1_i}$ and $x_{2_j}$ with appropriately redefined grid levels the feature sets $\xred$ and $\xadd$, respectively.  Note that in this case, each grid point now corresponds to the value of a vector of features.

It is also worth noting that \cite{MHJMLR} suggest comparing predictions generated with the full training set to not only those produced with the reduced set $\xred$, but also to those generated with $\xred$ and a permuted version of $\xadd$ in order to rule out the possibility that the ensemble is simply making use of additional noise.  The procedure we propose above avoids this potential confusion by utilizing the projection matrix $D$.

\section{Tests for additivity}
\label{sec:3}

We now demonstrate that this grid structure also allows for formal tests of additivity.

\subsection*{Tests for total additivity}
Again assume that our training set consists of only two features and that the response is observed according to $Y = F(X_1, X_2) + \epsilon$.  Tests for \emph{total} additivity assess whether the \emph{entire} underlying regression function $F$ is equal to, or at least well-approximated by, a sum of functions with disjoint domains.  When each function is univariate, this simply means that there are no interactions between any covariates but a more general case is also discussed below.  In the simple 2-dimensional case, the hypotheses of interest are
\begin{align}
\label{hyp:TotAdd1} &H_0:  \exists \; F_1, F_2 \mbox{  such that } F(x_1, x_2) = F_1(x_1) + F_2(x_2) \; \; \forall (x_1, x_2) \in \predpoints \\
&H_1:  F(x_1, x_2) \neq F_1(x_1) + F_2(x_2) \; \; \mbox{for some } (x_1, x_2) \in \predpoints \mbox{ for any } F_1, F_2. \notag
\end{align}

Again define a 2-dimensional grid of test points as in Figure \ref{fig:grid} so that each point in the grid has true value $F_{ij}$, predicted value $\hat{F}_{ij}$, and vectorized versions $V_F$ and $V_{\hat{F}}$.  Define $\bar{F}$ to be the mean of all predictions in the grid and define
\[
\hat{f}_{i \cdot} = \frac{1}{N_2} \sum_{j=1}^{N_2} \hat{F}_{ij}   \hspace{25mm} \mbox{ and } \hspace{25mm}  \hat{f}_{\cdot j} = \frac{1}{N_1} \sum_{i=1}^{N_1} \hat{F}_{ij}
\]
\noindent as the mean prediction at the $i^{th}$ level $x_{1_i}$ across all levels $x_{2_j}$, and the mean prediction at the $j^{th}$ level $x_{2_j}$ across all levels $x_{1_i}$, respectively.  If the features are additive, (i.e. under the null hypothesis) all points $(x_{1_i},x_{2_j})$ in the grid can be written as $F_{ij} = f_{i \cdot} + f_{\cdot j} - \mu$ where $\mu = \mathbb{E}\bar{F}$ is the true mean expected prediction across all points in the grid.  Thus, we may equivalently write the hypotheses in (\ref{hyp:TotAdd1}) as
\begin{align}
&H_0:  F_{ij} - f_{i \cdot} - f_{\cdot j} + \mu = 0 \; \; \mbox{for all } (x_1,x_2) \in \predpoints \notag \\
&H_1:  F_{ij} - f_{i \cdot} - f_{\cdot j} + \mu \neq 0 \; \; \mbox{for some } (x_1, x_2) \in \predpoints. \notag
\end{align}

The natural test statistic is then $\hat{F}_{ij} - \hat{f}_{i \cdot} - \hat{f}_{\cdot j} + \bar{F}$ which can be written as $D_2 V_{\hat{F}}$ where difference matrix is given by
\[
D_2 = I_{N} - \left( I_{N_1} \otimes \frac{1}{N_2} \bm{1}_{N_2 \times N_2} \right) - \left(\bm{1}_{N_1 \times N_1} \otimes \frac{1}{N_1} I_{N_2}\right) - \left(\frac{1}{N} \bm{1}_{N \times N}\right).
\]
Thinking of the $N_1$ and $N_2$ grid levels as factor levels of $X_1$ and $X_2$, we have $P = 1 + (N_1 - 1) + (N_2 - 1)$ degrees of freedom and $D_2$ has rank $N-P$.  As in Section 2, let $\Sigma$ denote the covariance of $V_{F}$ so that we can write $\Sigma_{D_2} = cov(D_2 V_{F}) = D_2 \Sigma D^{T}_{2}$ and use $(D_2 V_{\hat{F}})^{T} \hat{\Sigma}^{-1}_{D_2} D_2 V_{\hat{F}}\sim \chi_{N-P}^{2}$ as our test statistic.  Note that this testing procedure for total additivity is identical to the procedure for testing significance but in the final two steps we calculate an alternative difference matrix and test statistic.

This procedure also naturally extends to the case of $d$ features $X_1, ..., X_d$.  To test hypotheses of the form
\begin{align}
\label{hyp:FeatAdd} &H_0:  \exists \; F_1, ..., F_d \mbox{  s.t. } F(x_1, ..., x_d) = F_1(x_1) + \cdots + F_d(x_d) \; \; \forall (x_1, ...,x_d) \in \predpoints \\
&H_1:  F(x_1, ..., x_d) \neq F_1(x_1) + \cdots + F_d(x_d) \; \; \mbox{for some } (x_1, ..., x_d) \in \predpoints \mbox{ for any } F_1, ..., F_d \notag
\end{align}

\noindent we require a $d$-dimensional grid of test points so that given $N_i$ levels of each feature $X_i$, our grid contains a total of $N = \prod_{i=1}^{d} N_i$ test points.  Further, define

\[
\hat{f}_{\cdots j \cdots} = \frac{1}{N_1 \cdots N_{p-1} N_{p+1} \cdots N_{d}} \sum_{i_1 = 1}^{N_1} \cdots \sum_{i_{p-1} = 1}^{N_{p-1}} \sum_{i_{p+1} = 1}^{N_{p+1}} \cdots \sum_{i_d = 1}^{N_d} \hat{F}_{i_1 \cdots j \cdots i_d}
\]

\noindent to be the average prediction over all points in the grid at the $j^{th}$ level defined on the $p^{th}$ feature, $x_{p_j}$.  As in the 2-dimensional case, we can rewrite the hypotheses in (\ref{hyp:FeatAdd}) as
\begin{align}
&H_0:  F_{i_1 ... i_d} - f_{i_1 \cdots} - f_{\cdot i_2 \cdots} - \cdots - f_{\cdots i_d} + (d-1) \mu = 0 \; \; \mbox{for all } (x_{1}, ..., x_{d}) \in \predpoints \notag \\
&H_1:  F_{i_1 ... i_d} - f_{i_1 \cdots} - f_{\cdot i_2 \cdots} - \cdots - f_{\cdots i_d} + (d-1) \mu \neq 0 \; \; \mbox{for some } (x_{1}, ..., x_{d}) \in \predpoints \notag
\end{align}

\noindent and write $\hat{F}_{i_1 ... i_d} - \hat{f}_{i_1 \cdots} - \cdots - \hat{f}_{\cdots i_d} + (d-1) \bar{F}$ as $D_d V_{\hat{F}}$.  Again, we define $\Sigma$ to be the covariance of $V_{F}$ so that $\Sigma_{D_d} = cov(D_d V_{F}) = D_d \Sigma D^{T}_{d}$ and we can use  $(D_d V_{\hat{F}})^{T} \hat{\Sigma}^{-1}_{D_d} D_d V_{\hat{F}}\sim \chi_{N-P}^{2}$ as our test statistic, where $P = 1+(N_1-1) + \cdots + (N_d-1)$.

Importantly, the additive functions need not be univariate.  Define a (disjoint) partition of the feature space $\boldsymbol{S}_1, ..., \boldsymbol{S}_q$ so that $\cup_{i=1}^{q}\boldsymbol{S}_i = \{X_1, ..., X_d \}$.  We can test hypotheses of the form
\vspace{2mm}
\begin{align}
 &H_0:  \exists \; F_1, ..., F_q \mbox{  such that } F(\boldsymbol{s}_1, ..., \boldsymbol{s}_q) = F_1(\boldsymbol{s}_1) + \cdots + F_q(\boldsymbol{s}_q) \; \; \forall (\boldsymbol{s}_1, ...,\boldsymbol{s}_q) \in \predpoints \notag \\
&H_1:  F(\boldsymbol{s}_1, ..., \boldsymbol{s}_q) \neq F_1(\boldsymbol{s}_1) + \cdots + F_q(\boldsymbol{s}_q) \; \; \mbox{for some } (\boldsymbol{s}_1, ..., \boldsymbol{s}_q) \in \predpoints \mbox{ for any } F_1, ..., F_q \notag
\end{align}

\noindent in exactly the same fashion by appropriately defining levels of an $q$-dimensional grid of test points.

\subsection*{Tests for partial additivity}

We now handle the case where we are interested in testing only whether a proper subset of features contribute additively to the response.  Suppose that our training set consists of three features $X_1, X_2$, and $X_3$ and we are interested in testing
\begin{align}
\label{hyp:PartAdd} &H_0:  \exists \; F_1, F_2 \mbox{  s.t. } F(x_1, x_2, x_3) = F_1(x_1,x_3) + F_2(x_2,x_3) \; \; \forall (x_1, x_2,x_3) \in \predpoints \\
&H_1:  F(x_1, x_2, x_3) \neq F_1(x_1,x_3) + F_2(x_2,x_3) \mbox{ for some } (x_1, x_2,x_3) \in \predpoints \mbox{ for any } F_1, F_2. \notag
\end{align}

\noindent Rejecting this null hypothesis means that an interaction exists between $X_1$ and $X_2$ but implies nothing about potential interactions between $X_1$ and $X_3$ or between $X_2$ and $X_3$.  \cite{Hooker2004} uses the size of the deviation of $F$ from partial additivity as a means of identifying the bivariate and higher-order interactions required to reconstruct some percentage of the variation in the values of $F$.  This is also referred to as the Sobol index for the $X_1$, $X_2$ interaction \citep{Sobol2001}.  Define a 3-dimensional grid of test points with $N_1, N_2$, and $N_3$ levels of $X_1, X_2$ and $X_3$, respectively and continuing with the dot notation, define

\[
\hat{f}_{i \cdot k} = \frac{1}{N_2} \sum_{j=1}^{N_2} \hat{F}_{ijk}   \hspace{25mm} \mbox{ and } \hspace{25mm}  \hat{f}_{\cdot j k} = \frac{1}{N_1} \sum_{i=1}^{N_1} \hat{F}_{ijk}
\]

\noindent to be the average prediction over all levels of the feature $X_2$ in the grid at the $i^{th}$ and $k^{th}$ levels $x_{1_i}$ and $x_{3_k}$, and the average prediction over all levels of the feature $X_1$ in the grid at the $j^{th}$ and $k^{th}$ levels $x_{2_j}$ and $x_{3_k}$, respectively.  If there is no interaction between $X_1$ and $X_2$, then $F_{ijk} - f_{i \cdot k} - f_{\cdot j k} + f_{\cdot \cdot k} = 0$ at all levels $(x_{1_i},x_{2_j},x_{3_k})$ in the grid.  Thus, we can rewrite the hypotheses in (\ref{hyp:PartAdd}) as
\begin{align*}
&H_0:  F_{ijk} - f_{i \cdot k} - f_{\cdot j k} + f_{\cdot \cdot k} = 0 \; \; \forall (x_1, x_2, x_3) \in \predpoints \\
&H_1: F_{ijk} - f_{i \cdot k} - f_{\cdot j k} + f_{\cdot \cdot k} \neq 0 \; \; \mbox{for some } (x_1, x_2, x_3) \in \predpoints
\end{align*}

\noindent and use the empirical analogues of these parameters to conduct the testing procedure.  Once again, we can write $\hat{F}_{ijk} - \hat{f}_{i \cdot k} - \hat{f}_{\cdot j k} + \hat{f}_{\cdot \cdot k}$ as $D_3 V_{\hat{F}}$ for the appropriate difference matrix $D_3$.  Defining $\Sigma$ as the covariance of $V_{F}$, we can write $\Sigma_{D_3} = cov(D_3 V_{F}) = D_3 \Sigma D^{T}_{3}$ and use  $(D_3 V_{\hat{F}})^{T} \hat{\Sigma}^{-1}_{D_3} D_3 V_{\hat{F}}\sim \chi_{N-P}^{2}$ as our test statistic, where $N=N_1N_2N_3$.  Note that since we must now account for two-way interactions, we have $P = 1+(N_1-1)+(N_2-1)+(N_3-1)+(N_1-1)(N_3-1)+(N_2-1)(N_3-1)$ degrees of freedom and $D_3$ is of rank $N-P$.  As was the case in testing for total additivity, the testing procedure remains identical with the appropriate difference matrix and test statistic calculated in the final steps.

This same testing procedure can also be performed when our training set consists of $d$ features and we are interested in determining whether an interaction exists between $X_i$ and $X_j$.  Denote the set of all features except $X_i$ and $X_j$ as $\boldsymbol{X}_{-i,j}$ so that our hypotheses become
\begin{align*}
&H_0:  \exists \; F_i, F_j \mbox{  such that } F(x_1, ..., x_d) = F_i(x_i,\boldsymbol{x}_{-i,j}) + F_j(x_j,\boldsymbol{x}_{-i,j}) \; \; \forall (x_1, ..., x_d) \in \predpoints \\
&H_1:  F(x_1, ..., x_d) \neq F_i(x_i,\boldsymbol{x}_{-i,j}) + F_j(x_j,\boldsymbol{x}_{-i,j}) \mbox{ for some } (x_1, ..., x_d) \in \predpoints \mbox{ for any } F_i, F_j.
\end{align*}

Now, instead of the third dimension of the grid containing levels of the single feature $X_3$, these are now vector levels $\bm{x}_{-i,j}$ and the testing procedure remains identical.  Likewise, $X_i$ and $X_j$ may be treated as vectors of features by redefining the grid levels as levels of the appropriate vector. \\

\vspace{1mm}

\noindent {\bf Remark:  } The testing procedures above as well as those defined in Section 2 were derived assuming equal weight is placed on each point in the test grid.  In some cases, it may be advantageous to instead differentially weight grid points, for example based on the local density of observations.  This alternative approach based on minimizing a weighted sum of squared errors is outlined in Appendix A.  For a more thorough review of when such an alternative may be preferred, we refer the reader to \cite{Hooker2007}.

\section{Random Projections}
\label{sec:4}
The above procedures require estimating a covariance matrix of size proportional to the number of points in the test grid.  However, estimating the variance parameters with too small an ensemble can result in a significant overestimate of the variance, thereby substantially reducing the power of our testing procedures; see \cite{MHJMLR} for a more complete discussion.  Thus, in situations where large grids and/or complex additive forms are of interest, it may become computationally infeasible to directly obtain an accurate covariance estimate.  In light of this, we further extend our above procedures to make use of random projections.


Random projections have a long-established history as a dimension-reduction method.  The Johnson-Lindenstrauss Lemma \citep{Johnson1984} provides that orthogonal projections from high dimensional spaces into lower dimensional spaces approximately preserve the distances between the projected elements.  \cite{Lopes2011} and \cite{Srivastava2015} leverage this result to produce a high-dimensional extension of Hotelling's classic $T^2$ test to the $p>n$ case.  Specifically, given two multivariate samples $X_{n_1 \times p}$ and $Y_{n_2 \times p}$, the data is projected via a random projection matrix $R$ into a reduced dimension $r < n,p$ where an analogous testing procedure can be well-defined.  In the latter work, the authors denote this test RAPTT ({\bf Ra}ndom {\bf P}rojection {\bf T}-{\bf T}est) and for each projection matrix $R_i$, the projected test statistic and p-value are given by

\[
T_{R_i}^{2} = \frac{1}{n_{1}^{-1}+n_{2}^{-1}} (\bar{X}-\bar{Y})' R_i (R_i' S R_i)^{-1} R^{'}_i (\bar{X}-\bar{Y})
\]

\noindent and

\[
\theta_{R_i} = 1 - F_{r,n-r+1} \left( \frac{n-r+1}{r} \frac{T^{2}_{R_i}}{n} \right)
\]


\noindent where $n = n_1 + n_2 - 2$, $S$ is the (pooled) sample covariance matrix, and $F_{a,b}$ denotes the $F$-distribution with numerator and denominator degrees of freedom $a$ and $b$, respectively.  RAPTT proceeds by sampling $M$ random projection matrices $R_1, ..., R_M$ thereby obtaining a total of $M$ of the test statistics and p-values defined above.  The final test statistic in the procedure with level $\alpha$ is defined as the average across the $M$ p-values, $\theta = \frac{1}{M} \sum_{i=1}^{M} \theta_{R_i}$ and the null hypothesis of equal means is rejected whenever $\theta < u_{\alpha}$ where $u_{\alpha}$ is chosen such that $P \left[ \theta < u_{\alpha} \Big| H_0 \right] = \alpha$.  

In our context, we consider a training set of size $n$, an ensemble consisting of $m$ trees, each of which is built with a subsample of size $k$, and we are interested in predicting at $N$ total test points.  Recall from Section \ref{sec:2} that the simplest form of test statistic that can be used to evaluate variable importance is given by $\hat{\mu}_{N}^{T} \hat{\Sigma}^{-1} \hat{\mu}_N \sim \chi_{N}^{2}$ where $\hat{\mu}$ is the vector of ensemble predictions, and $\hat{\Sigma}$ is the corresponding covariance matrix estimate.  Given $m$ predictions at each of $N$ locations, we can think of our data as an $m \times N$ matrix so that for a set of $M$ random projection matrices $R_1, ..., R_M$ and reduced dimension $r < m,N$, we can write each projected test statistic as

\begin{equation}
\label{tstat:origproj}
T_{R_i} = \hat{\mu}_{N}^{T} R_i (R^{T}_{i} \hat{\Sigma} R_i)^{-1} R^{T}_{i} \hat{\mu}_N \sim \chi^{2}_{r}.
\end{equation}



\noindent The grid structure can also be incorporated in a straightforward manner.  Though we utilize a difference matrix $D$ to project into the space of additive models, so long as the elements of the $R_i$ are independently generated continuous random variables, the overall projection has rank $r$ with probability 1.  The original test statistic is given by $(D V_{\hat{F}})^{T} \hat{\Sigma}^{-1}_{D} D V_{\hat{F}}$  where $\Sigma_{D} = cov(D V_{F}) = D \Sigma D^{T}$ and so the test statistic and p-value incorporating a random projection $R_i$ become

\[
T_{R_i} = (D V_{\hat{F}})^{T} R_i (R^{T}_{i} \hat{\Sigma}_{D} R_i)^{-1} R^{T}_{i} (D V_{\hat{F}}) \sim \chi^{2}_{r}
\]

\noindent and

\[
\theta_{i} = 1 - \Phi^{2}_{r}(T_{R_i})
\]

\noindent respectively, where $r < N-P$ and $\Phi^{2}_{r}$ denotes the cdf of the $\chi^{2}_{r}$.  For $M$ replicates of this randomized testing procedure, we can define our final test statistic as $\theta = \frac{1}{M} \sum_{i=1}^{M} \theta_i$ in the same fashion as RAPTT, where we reject $H_0$ whenever $\theta < u_{\alpha}$ and $u_{\alpha}$ is chosen such that $P \left[ \theta < u_{\alpha} \Big| H_0 \right] = \alpha$.

%




\subsection{Defining the Testing Parameters}
The procedures developed in the preceding sections require a number of user-specified parameters.  First, as noted in \cite{Srivastava2015}, the choice of reduced dimension $r$ is an important consideration that can influence the power of projection-based testing procedures.  In our case, the covariance parameters are difficult to estimate accurately on large grids and thus, though the procedure is well-defined for $1 \leq r < m$, this practical restriction necessitates a relatively small projected dimension $r$.  In many cases, we see a significant drop in power when testing on grids consisting of more than approximately 30 points, so choosing $5 \leq r \leq 15$ should be reasonable and computationally feasible in most situations.  Further, note that because $r$ is small, little dependence remains between the resulting p-values.  Under the null hypothesis, each p-value is uniformly distributed on $[0,1]$ and the mean of independent standard uniform random variables follows a Bates distribution, so the final cutoff $u_{\alpha}$ can be well approximated by the $\alpha$ quantile of this distribution.




The ideal method of sampling the random projection matrices is of less concern; \cite{Srivastava2015} show that any semi-orthogonal matrix $R$ with elements generated from a continuous distribution with finite second moment satisfies the necessary conditions to perform the projection-based tests.  For our situation, we recommend generating such matrices by sampling individual elements from a standard normal distribution, orthogonalizing via a process such as Gram-Schmidt, and selecting the appropriate submatrix.  Such a procedure is straightforward and can be implemented in most software packages.

Algorithm 1 makes the random-projection-based testing procedure explicit, using the internal variance estimation procedure proposed in \cite{MHJMLR}.  For a particular query point of interest $\bm{x}$, the asymptotic variance of the prediction is given by
\[
\frac{k^{2}}{m \, \alpha} \zeta_{1,k} + \frac{1}{m}\zeta_{k,k}
\]

\noindent where $\zeta_{1,k} = \mbox{var}\left( E \left( T_i(\bm{x}) \big| \tilde{\bm{x}} \right) \right)$ represents the variance between tree-based predictions $T_i(\bm{x})$ at $\bm{x}$ given a single common training point $\tilde{\bm{x}}$, $\zeta_{k,k} = \mbox{var} \left( T_i(\bm{x})  \right)$ denotes the between-tree variance, and $\alpha = \lim n/m$.  The algorithm makes use of the parameters $n_{\tilde{\bm{x}}}$ and $n_{MC}$ in order to structure the ensemble in such a fashion so as to readily post-compute consistent estimates of $\zeta_{1,k}$ and $\zeta_{k,k}$.  The parameter $n_{\tilde{\bm{x}}}$ corresponds to the number of conditional expectation estimates $E \left( T_i(\bm{x}) \big| \tilde{\bm{x}} \right)$ computed in the definition of $\zeta_{1,k}$ and $n_{MC}$ is the number of Monte Carlo samples used to estimate each conditional expectation so that in this case, $m = n_{\tilde{\bm{x}}} \times n_{MC}$.  Though the ensemble need not be constructed in such a fashion, the internal estimation procedure allows us to easily select a small projected dimension $r$ and also allows for the covariance estimates to be computed at no additional cost to the original ensemble.

\begin{algorithm}
\hspace{1mm} Compute difference matrix $D$ \\
\hspace{1mm} Select reduced dimension $r$ \\
\hspace{1mm} Generate random projection matrices $R_1, ..., R_M$ \\
\hspace{1mm} {\bf for} $i$ in 1 to $n_{\tilde{\bm{x}}}$

\hspace{8mm} Select initial fixed point $\tilde{\bm{x}}^{(i)}$

\hspace{8mm} {\bf for} $j$ in 1 to $n_{MC}$

\hspace{12mm} Select subsample $\mathcal{S}_{\tilde{\bm{x}}^{(i)},j}$ of size $k_n$ from training set that includes $\tilde{\bm{x}}^{(i)}$

\hspace{12mm} Build tree using subsample $\mathcal{S}_{\tilde{\bm{x}}^{(i)},j}$

\hspace{12mm} Use tree to predict at each of the $N$ grid points to obtain $\hat{V}_j$

\hspace{12mm} Apply each projection to $\hat{V}$ to obtain $\hat{W}_{j,c} = (D\hat{V}_j)^{T}R_c$
	
\hspace{8mm} {\bf end for}

\hspace{8mm} Record average of $\hat{W}_{j,c}$ over $j$ for each projection

\hspace{1mm} {\bf end for}

\hspace{1mm} Compute variance of each of the $n_{\tilde{\bm{x}}}$ averages to estimate each $\zeta_{1,k_n}$ \\
\hspace{1mm} Compute variance of all predictions from each projection to estimate each $\zeta_{k_n,k_n}$ \\
\hspace{1mm} Compute mean of all predictions from each projection to estimate each $\theta_{k_n}$ \\
\hspace{1mm} Compute each p-value $\theta_1, ..., \theta_M$ by comparing to $\chi_{r}^{2}$ \\
\hspace{1mm} Record average p-value $\theta$ and compare to Bates $\alpha$ quantile
\caption{Random Projection Testing Procedure}
\label{algo:largep}
\end{algorithm}

Finally, the levels of the test grid are an important consideration.  As with all supervised learning procedures, these test points should be concentrated near the observed data so as to minimize the effects of extrapolation.  However, with tree-based procedures, choosing grid points that appear in the original sample can also be problematic.  Because the trees in random forests are grown to near full-depth without pruning, predictions made arbitrarily close to points in the training sample can suffer from overfitting and as a result, create the artificial appearance of interactions.  Lastly, because predictions are based on localized averaging, grid points should be selected away from the boundary of the feature space to avoid edge effects.  In most situations, uniformly spaced grid points in the interior of the feature space should produce tests with high power that preserve the level of the test.

\section{Simulations}
\label{sec:5}

We now provide simulations to investigate the power of our proposed testing procedures.  Suppose first that we have two features $X_1$ and $X_2$ and that our responses are generated according to $Y = X_1 + X_2 + \beta X_1X_2 + \epsilon$ where we set $\beta=0$ to assess $\alpha$-level and $\beta=1$ to evaluate power with $\epsilon \sim \mathcal{N}(0,0.05^2)$.  We first test for total additivity on 1000 datasets when $\beta=0$ and 1000 datasets where $\beta=1$, taking our empirical $\alpha$-level as the proportion of tests that incorrectly reject the null hypothesis (when $\beta=0$) and our estimate of power as the proportion of tests that correctly reject the null hypothesis (when $\beta=1$).  For reference, we also built 1000 linear regression models using the traditional t-test to determine whether the interaction is significant and recorded the empirical $\alpha$-level and power of this testing procedure.  This was repeated for data sets of size of 250, 500, and 1000 using subsample sizes of 30, 50, and 75 respectively and the results are shown in Table \ref{table:1}.   The test grid was selected as a $4 \times 4$ grid with levels 0.2, 0.4, 0.6, and 0.8.  In each case, our test for total additivity using a subbagged ensemble performed nearly exactly as well as the traditional t-test.

\begin{table}[t]
\label{table:1}
\begin{center}
\begin{tabular}{|r|c|c|c|}
\hline
\multicolumn{1}{|c|}{\textbf{Method}} & \textbf{$\boldsymbol{n}$}            & \textbf{$\boldsymbol{\alpha}$-level} & \textbf{Power} \\ \hline
Linear Model                          & \multirow{2}{*}{250}  & 0.056            & 1.000          \\ \cline{1-1} \cline{3-4}
Subbagged Ensemble                            &                       & 0.065            & 0.954          \\ \hline
Linear Model                          & \multirow{2}{*}{500}  & 0.048            & 1.000          \\ \cline{1-1} \cline{3-4}
Subbagged Ensemble                            &                       & 0.047            & 0.998          \\ \hline
Linear Model                          & \multirow{2}{*}{1000} & 0.046            & 1.000          \\ \cline{1-1} \cline{3-4}
Subbagged Ensemble                            &                       & 0.020            & 0.999          \\ \hline
\end{tabular}
\end{center}
\caption{\label{table:1} Empirical $\alpha$-levels and power for the linear model example.}
\end{table}

We also selected a number of more complex regression functions that have been used in previous publications related to testing additivity, such as \cite{De2004} and \cite{Barry1993}, to further investigate $\alpha$-level and power.  Each estimate is the result of 1000 simulations with a sample size of 500, subsample size of 50, and a $4 \times 4$ test grid (with levels 0.2, 0.4, 0.6, and 0.8) in the 2-dimensional tests for total additivity and a $3 \times 3 \times 3$ grid (with levels 0.3, 0.5, and 0.7) in the 3-dimensional tests for total and partial additivity.  In each case the features were selected uniformly at random from $[0,1]$, the responses generated according to $Y = F(\bm{X}) + \epsilon$ with $\epsilon \sim \mathcal{N}(0,\sigma^2)$ with $\sigma$ chosen to take values $0.05$, $0.25$ or $0.5$, and the covariance estimated via the internal estimation procedure.  The results are shown in Table \ref{table:2} where the first line for each model gives the rejection probability for the tests defined here.  Note that even though the response in the first two models does not depend on $X_2$, this additional feature was still included in the training sets and the same test for total additivity was performed.  In each case, we see that our false rejection rate is very conservative and we also maintain high power.  Note that in each of these simulations, the variance estimation parameters were selected as $n_{\tilde{\bm{x}}}=50$ and $n_{MC}=250$.  These parameters assignments are smaller than those chosen in \cite{MHJMLR} and the authors note that these smaller ensemble sizes often lead to an overestimate of the variance thus resulting in the conservative test results (low $\alpha$-levels) seen in Table 2.

\begin{table}[t]
\begin{adjustbox}{width=\textwidth}
\footnotesize
\begin{tabular}{|l|c|ccc|l|c|ccc|}
\hline
\multicolumn{1}{|c|}{\textbf{Model}}             & \textbf{Test}    & \multicolumn{3}{c|}{\textbf{Noise s.d.}} & \multicolumn{1}{c|}{\textbf{\textbf{Model}}}                                & \textbf{Test}    & \multicolumn{3}{c|}{\textbf{Noise s.d.}} \\
\multicolumn{1}{|c|}{}                             &                    & $0.5$        & $0.25$       & $0.05$       & \multicolumn{1}{c|}{}                                                         &                    & $0.5$        & $0.25$       & $0.05$       \\ \hline
\multirow{2}{*}{(a) $x_1$}                         & \multirow{2}{*}{T} & 0.009        & 0.007        & 0.000        & \multirow{2}{*}{(h) \hspace{0.5mm} $x_1x_2$}                                                 & \multirow{2}{*}{T} & 0.085        & 0.702        & 1.000        \\
                                                   &                    & 0.025        & 0.031        & 0.000        &                                                                               &                    & 0.305        & 0.927        & 1.000        \\ \hline
\multirow{2}{*}{(b) $e^{x_1}$}                     & \multirow{2}{*}{T} & 0.002        & 0.000        & 0.000        & \multirow{2}{*}{(i) \hspace{0.5mm} $x_1x_2x_3$}                                              & \multirow{2}{*}{P} & 0.001        & 0.007        & 0.948        \\
                                                   &                    & 0.028        & 0.011        & 0.000        &                                                                               &                    & 0.002        & 0.028        & 0.998        \\ \hline
\multirow{2}{*}{(c) $e^{x_1} + \sin(\pi x_2)$}     & \multirow{2}{*}{T} & 0.008        & 0.008        & 0.007        & \multirow{2}{*}{(j) $\frac{\exp(5(x_1 + x_2))}{1 + \exp(5(x_1 + x_2))} - 1$} & \multirow{2}{*}{T} & 0.006        & 0.029        & 0.948        \\
                                                   &                    & 0.045        & 0.060        & 0.059        &                                                                               &                    & 0.021        & 0.089        & 0.999        \\ \hline
\multirow{2}{*}{(d) $x_1+x_2+x_3$}                 & \multirow{2}{*}{T} & 0.002        & 0.003        & 0.001        & \multirow{2}{*}{(k) $\frac{1 + \sin(2 \pi (x_1 + x_2))}{2}$}                 & \multirow{2}{*}{T} & 1.000        & 1.000        & 1.000        \\
                                                   &                    & 0.000        & 0.001        & 0.001        &                                                                               &                    & 1.000        & 1.000        & 1.000        \\ \hline
\multirow{2}{*}{(e) $e^{x_1} + e^{x_2} + e^{x_3}$} & \multirow{2}{*}{T} & 0.003        & 0.007        & 0.007        & \multirow{2}{*}{(l) $\frac{1 + \sin(2 \pi (x_1 + x_2 + x_3))}{2}$}           & \multirow{2}{*}{P} & 0.158        & 0.874        & 1.000        \\
                                                   &                    & 0.005        & 0.019        & 0.012        &                                                                               &                    & 0.011        & 0.222        & 0.959        \\ \hline
\multirow{2}{*}{(f) $x_1x_3+x_2x_3$}               & \multirow{2}{*}{P} & 0.000        & 0.000        & 0.002        &\hspace{-1.5mm} (m) $64 (x_1 x_2)^3$                                                         & \multirow{2}{*}{T} & 0.907        & 1.000        & 1.000        \\
                                                   &                    & 0.000        & 0.001        & 0.008        & $\hspace{1cm}  (1 - x_1 x_2)^3$                                             &                    & 0.987        & 1.000        & 1.000        \\ \hline
\multirow{2}{*}{(g) $e^{x_1x_3}+e^{x_2x_3}$}       & \multirow{2}{*}{P} & 0.000        & 0.001        & 0.014        &\hspace{-1.25mm} (n) $64 (x_1 x_2 x_3)^3$                                                     & \multirow{2}{*}{P} & 0.051        & 0.722        & 0.999        \\
                                                   &                    & 0.000        & 0.006        & 0.066        & $\hspace{1cm}  (1 - x_1 x_2 x_3)^3$                                         
&                    & 0.136        & 0.898        & 1.000        \\ \cline{1-10}
\end{tabular}
\end{adjustbox}
\caption{\label{table:2} Empirical $\alpha$-level and power for a variety of underlying regression functions with noise levels of different standard deviations. Tests are either for Total (T) or Partial (P) additivity; for each model, the top result represents the power for a test without random projections, the bottom for the
test employing random projections.  The lettered labels beside each model are for comparison purposes to Figure 2.  }	
\end{table}

\begin{figure}[h]
\begin{center}
\includegraphics[width=0.95\textwidth]{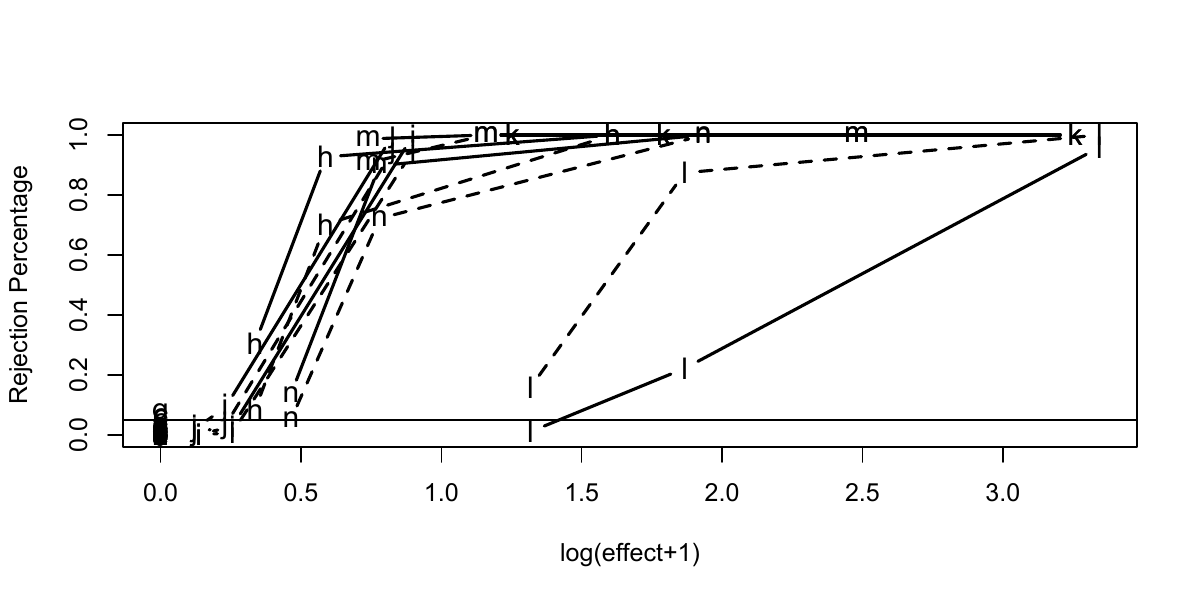}
\end{center}
\caption{Graphical representation of the proportion of rejected tests out of 1000 trials corresponding to Table \ref{table:2}. The $x$-axis plots the non-centrality parameter for the non-random projection test for each case. Lines connect tests of the same model at different model variances; solid lines represent tests with random projections, dashed lines without.  } \label{fig:modelPower}
\end{figure}

Next, we repeated these simulations on the same functions, this time employing our tests that utilize random projections.  In the 2-dimensional tests for total additivity, we use a $10 \times 10$ grid so that $N=100$ and in the 3-dimensional tests for total additivity and the tests for partial additivity, we use a $5 \times 5 \times 5$ grid so that $N=125$.  The results are shown in Table \ref{table:2} in the second row for each model.  Note that in these tests, we maintain a reasonable type 1 error rate but achieve significantly more power due to the finer resolution of the test grid. These results are also presented graphically in Figure \ref{fig:modelPower} where we can see that the tests utilizing random projections tend to have higher power. The only exception to this is model (l) with $y = 0.5(1 + \sin(2 \pi (x_1 + x_2 + x_3))) + \epsilon$ where the complexity of the response surface and the choice of evaluation points likely affected the outcome.

\begin{figure}[h]
\begin{center}
\begin{tabular}{ccc}
Total Additivity & Partial Additivity & Variable Importance \\
\includegraphics[height=4.5cm]{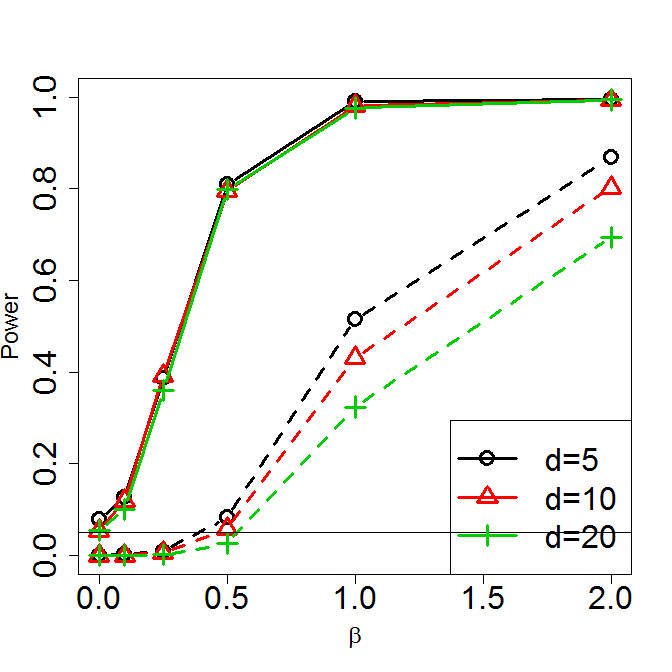} &
\includegraphics[height=4.5cm]{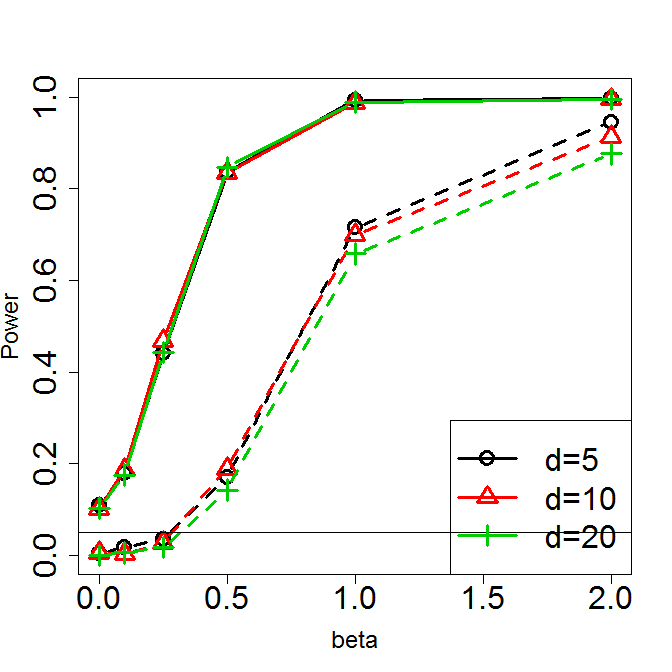} &
\includegraphics[height=4.5cm]{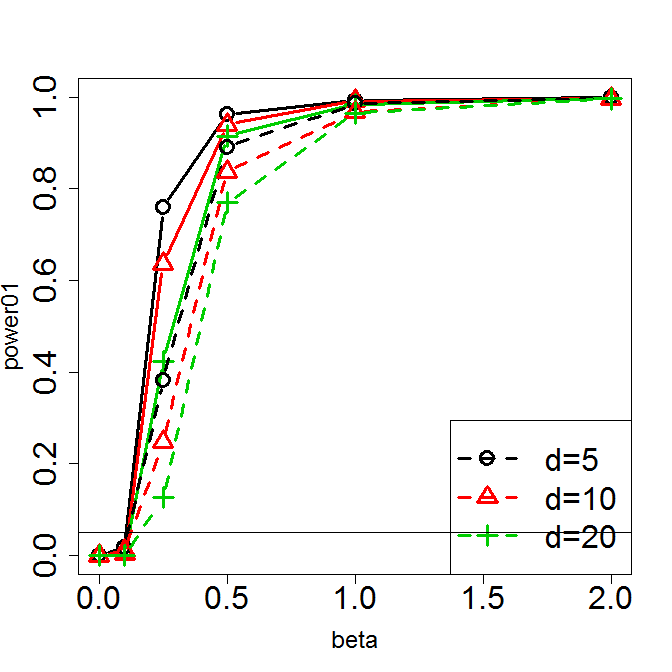}
\end{tabular}
\end{center}
\caption{Results of a simulated power experiment. The left panel provides the power of a test of total additivity between two covariates with strength governed by $\beta$ in the presence of $d$ additional nuisance covariates. The middle panel repeats this procedure with tests of partial additivity.  The right panel tests the importance of a  three-dimensional set of covariates in the presence of an additional $d+3$ covariates. In both cases, responses were generated with Gaussian errors  with standard deviation 0.1 (solid lines) or 1.0 (dashed lines). Here we observe sensitivity to error variance, but a relatively small impact of nuisance covariate dimension.} \label{fig:power}
\end{figure}

The computational effort required to perform these tests is proportional to the dimension and overall size of the chosen grid.  That is, tests of a particular form may be carried out at little additional computational cost for larger dimensions of the covariate space.  To demonstrate this point, we first examine a test of total additivity. Here we again employ the model $Y = X_{1} + X_{2} + \beta X_1 X_2  + \epsilon$ where covariates are sampled uniformly from $[-1, 1]^{d+2}$, $d$ takes values $5$, $10$ and $20$, and $\epsilon$ is chosen to be either $N(0,0.01)$ or $N(0,1)$.  Here $d$ represents the dimension of nuisance covariates and $\beta$ is taken to be one of $0$, $0.1$, $0.25$, $0.5$, $1$, $2$, giving the strength of the interaction.  For each combination of $\beta$ and $d$, we employ a $5 \times 5$ grid with points selected uniformly in [-0.6, 0.6] and utilize 1000 random projections with $r=5$.  Selecting interior grid points in [-0.6, 0.6] helps avoid the potential edge effects common in tree-based methods when predicting near the boundary of the feature space.  For each of these settings, we generated 1000 datasets of 500 observations from which we obtained a random forest with subsamples of size 50 and conducted tests of total and partial additivity.  The results of this simulation are given in left two panels of Figure \ref{fig:power} where we see that these tests achieve approximately the correct $\alpha$ level at $\beta = 0$, but quickly produce high power. We observe an expected drop in power with increasing error variance, but relative insensitivity to nuisance dimensions. \\


We next extend this experiment to testing the importance of a {\em group} of variables. Here we employ the model
\[
Y =  \beta (X_{1}+X_{2}+X_{3}) + X_{4}+X_{5}+X_{6} + \epsilon
\]
under the same data generation scheme. Here we test the joint significance of $(X_1,X_2,X_3)$ while also including further signal from $(X_4,X_5,X_6)$.  As above, we used $d = 5$, 10 or 20 additional nuisance covariates and $\beta$ is taken to be one of $0$, $0.1$, $0.25$, $0.5$, $1$, $2$, giving the strength of the signal from the first three covariates.  The $\epsilon_i$ are again normal with standard deviation 0.1 or 1. The right panel in Figure \ref{fig:power} shows the empirical power of the test of importance for the vector $(X_1,X_2,X_3)$  based on 1000 simulations of datasets of size 500 and subsamples of size 50.  For each combination of $\beta$ and $d$, we employ a $5 \times 5 \times 5$ grid with points selected uniformly in [-0.6, 0.6] and utilize 1000 random projections with $r=5$.  We observe approximately the correct $\alpha$-level at $\beta = 0$ with power increasing with $\beta$, resulting in power of approximately 0.8 at $\beta = 0.5$. 
These results are in agreement with \cite{biau12} in which it is suggested that random forests are largely able ignore nuisance covariates with power decreasing only marginally with larger nuisance dimension $d$.
\section{Real data}
\label{sec:6}

We now demonstrate our testing procedures on a dataset provided by a team of ornithologists at the Cornell University Lab of Ornithology.  This dataset was compiled in an effort to determine how pollution levels affect the change in Wood Thursh population.  The data consists of 3 pollutant features, mercury deposition ($md$), acid deposition ($ad$), and soil PH level ($sph$) as well as 2 non-pollutant features, elevation ($elev$) and abundance ($ab$).  We begin our analysis by testing whether the pollutant and non-pollutant features are additive:
\begin{align}
&H_0:  F(md,ad,sph,elev,ab) = F_P(md,ad,sph) + F_{NP}(elev,ab). \label{hyp:ebird1}
\end{align}

In this case we have two feature sets, $\bm{X}_1 = (md,ad,sph)$ and $\bm{X}_2 = (elev,ab)$ and we performed a test for total additivity using 4 levels of each set -- the 0.20, 0.40, 0.60, and 0.80 quantiles of each feature -- for a total of 16 test points.  Our test statistic was 52.30, larger than the critical value, the 0.95 quantile of the $\chi^{2}_{9}$, of 16.92 so we reject the null hypothesis in (\ref{hyp:ebird1}) and conclude that an interaction exists between the pollutant and non-pollutant features.  This result was confirmed by our random projection test, which consisted of 1000 random projections to a dimension of $r=5$ using a $10 \times 10$ test grid.  In this case, the final averaged p-value was only 0.0043, far below the critical value of 0.485.

Next, we investigated how the pollutants contributed to the response.  Based on preliminary investigations, ebird researchers suspected an interaction between mercury and acid deposition ($md$ and $ad$) but were unsure of the relationship between soil PH ($sph$) and $md$ and $ad$.  In performing these tests for partial additivity, our test grid consisted of 3 points for each feature set, the 0.30, 0.50, and 0.70 quantiles of each feature for a total of 27 test points and a critical value, the 0.95 quantile of the $\chi_{12}^{2}$, of 21.03.  Our test for partial additivity between $md$ and $ad$,
\vspace{1mm}
\begin{align}
&H_0:  F(md,ad,sph,elev,ab) = F_1(md,sph,elev,ab) + F_2(ad,sph,elev,ab), \notag
\end{align}

\noindent yielded a significant result with a test statistic of 41.00 so our test supports the belief that an interaction exists between $md$ and $ad$.   Again, this result was supported by our random projection test, which consisted of 1000 random projections to a dimension of $r=5$ using a $5 \times 5 \times 5$ test grid, for a total of 125 test points.  The final averaged p-value was only 0.0064, far below the critical value of 0.485.

Our test for partial additivity between $sph$ and the vector $(md,ad)$
\vspace{1mm}
 \begin{align}
&H_0:  F(md,ad,sph,elev,ab) = F_1(md,ad,elev,ab) + F_2(sph,elev,ab) \notag
\end{align}

\noindent yielded a test statistic of 36.43, above the critical value of 21.03, so once again we reject the null hypothesis and conclude that an interaction exists between $sph$ and $(md,ad)$.  This result was again supported by the random projection test based on 1000 random projections to a dimension of $r=5$ using a $5 \times 5 \times 5$ test grid.  We find a final averaged p-value of 0.225, which, though larger than in the previous tests, is still far below the critical value of 0.485.

\section{Discussion}

This work harnesses desirable asymptotic properties of subsampled ensemble learners to develop formal hypothesis tests for additivity in random forests and suggests that traditional scientific and statistical questions need not be seen as a sacrifice of less interpretable learning procedures.  Our tests require the definition of a reasonably sized test grid in order achieve reasonably accurate covariance estimates while preserving power.  When larger grids or more complex additive forms are required, we appeal to random projections and demonstrate that our tests still maintain very high power.

Many of the above demonstrations employed a version of random forests in which each covariate remains eligible at each split (subbagged ensembles), though we point out that the theory established in previous work such as \cite{MHJMLR}, \cite{Wager2015}, and \cite{Scornet2015} allows for most general subsampled random forests implementations, or in fact any ensemble-type learner that conforms to the regularity conditions to be used.  We caution however that  the predictive improvement often seen with random forests is generally attributed to the increased independence between trees and thus should be expected to be less dramatic in these cases where subsamples are used in lieu of the traditional bootstrap samples.

Finally, it is important to note that the particular additive forms for which the testing procedures were developed were chosen only because of their scientific utility.  Testing procedures for alternative additive forms can be developed in a similar manner by establishing appropriate model parameters from an ANOVA set-up and defining the difference matrix $D$ accordingly.  These methods can also be extended to provide formal statistical guarantees for the screening procedures described in \cite{Hooker2004}.

\bibliographystyle{apalike}

\section*{Appendix}
\begin{appendix}
\section{The generalized approach}
\label{app:A}


The testing procedures developed in Sections 2 and 3 were derived by choosing the model parameters that minimized the sum of squared error (SSE) with equal weight placed on each point in the test grid.  Instead, we may wish to differentially weight points on the grid. For example, in the above tests for partial additivity, we can select $\hat{F}_1$ and $\hat{F}_2$ to minimize the weighted SSE

\[
WSSE = \sum_{i,j,k} w_{i,j,k} \left(F(x_{1_i},x_{2_j},x_{3_k}) - F_{1}(x_{1_i},x_{3_k}) - F_{2}(x_{2_j},x_{3_k}) \right)^2
\]

\noindent where $x_1, x_2, x_3$ can be taken as individual features or interpreted more generally as vectors of features and the weights $w_{ijk}$ are specified by the user.  \cite{Hooker2007} recommends basing such weights on an approximation to the density of observations near $(x_{1_i},x_{2_j},x_{3_k})$.  This procedure takes the form of a weighted ANOVA.  In particular, define $\vec{F}$ to be the $N_1 N_3 + N_2 N_3$ vector concatenating the $\hat{F}_{1}(x_1,x_3)$ and $\hat{F}_{2}(x_2,x_3)$ and as in the previous sections let $V_{\hat{F}}$ be the vector containing the $\hat{F}_{ijk}$.  Further, let $Z$ be the $N \times \left(N_1 N_3 + N_2 N_3\right)$ matrix defined so that $Z \vec{F}$ produces the corresponding $\hat{F}_{1}(x_1,x_3) + \hat{F}_{2}(x_2,x_3)$ and let $W$ be a diagonal matrix containing the weights.  Then we can write

\[
WSSE= (V_{\hat{F}} - Z \vec{F})^T W (V_{\hat{F}} - Z \vec{F})
\]

\noindent and we know that the solution $\vec{F}$ that minimizes this weighted SSE is given by

\[
\vec{F} = (Z^T W Z)^{-1} Z^T W V_{\hat{F}}
\]

\noindent so that under the null hypothesis

\[
V_{\hat{F}} - Z \vec{F} = (I - Z(Z^T W Z)^{-1} Z^T W) V_{\hat{F}}
\]

\noindent has mean 0.  Further, letting $\Sigma$ denote the covariance of $V_{F}$, the variance of $V_{F} - Z \vec{F}$ is given by

\[
C = (I - Z(Z^T W Z)^{-1} Z^T W) \Sigma (I - Z(Z^T W Z)^{-1} Z^T W)^T
\]

\noindent so that

\[
\left[ (I - Z(Z^T W Z)^{-1} Z^T W) V_{\hat{F}} \right]^T \hat{C}^{-1} \left[ (I - Z(Z^T W Z)^{-1} Z^T W) V_{\hat{F}} \right]
\]

\noindent has a $\chi^2_{N-P}$ distribution, where $P$ remains as defined in the standard procedures developed in Sections 2 and 3.  For equal weighting ($W$ given by the identity matrix), these calculations reduce to the averages employed above, and for the sake of simplicity we have restricted ourselves to this choice.  Note also that this generalized WLS approach can be applied to more general forms of additivity as well as those tests for total additivity developed in the previous section.

\end{appendix}

\end{document}